\documentclass[12pt]{article}
\pdfoutput=1
\usepackage{scicite}
\usepackage{times}
\usepackage{subfig}
\usepackage{epsfig}
\usepackage{amsmath}
\usepackage{hyperref}

\def\beqa#1\eeqa{\begin{eqnarray}#1\end{eqnarray}}

\newenvironment{sciabstract}{%
\begin{quote} \bf}
{\end{quote}}

\newcounter{lastnote}
\newenvironment{scilastnote}{%
\setcounter{lastnote}{\value{enumiv}}%
\addtocounter{lastnote}{+1}%
\begin{list}%
{\arabic{lastnote}.}
{\setlength{\leftmargin}{.22in}}
{\setlength{\labelsep}{.5em}}}
{\end{list}}

\makeatletter
\def\fnum@figure{Fig. \thefigure}
\makeatother

\title{Improving neural networks by preventing co-adaptation of feature detectors}

\author{G. E. Hinton$^\ast$, N. Srivastava, A. Krizhevsky, I. Sutskever and R. R.  Salakhutdinov\\
\normalsize{Department of Computer Science, University of Toronto,}\\
\normalsize{6 King's College Rd, Toronto, Ontario M5S 3G4, Canada}\\
\\
\normalsize{$^\ast$To whom correspondence should be addressed; E-mail:  hinton@cs.toronto.edu}
}

\date{}

\begin{document} 



\maketitle 

\begin{sciabstract}
When a large feedforward neural network is trained on a small training set, it typically
performs poorly on held-out test data.  This ``overfitting'' is greatly reduced by
randomly omitting half of the feature detectors on each training case. This prevents
complex co-adaptations in which a feature detector is only helpful in the context of
several other specific feature detectors.  Instead, each neuron learns to detect a feature
that is generally helpful for producing the correct answer given the combinatorially large
variety of internal contexts in which it must operate. Random ``dropout'' gives big
improvements on many benchmark tasks and sets new records for speech and object
recognition.
\end{sciabstract}

A feedforward, artificial neural network uses layers of non-linear ``hidden''
units between its inputs and its outputs. By adapting the weights on the incoming
connections of these hidden units it learns feature detectors that enable it to predict
the correct output when given an input vector \cite{RHW}.  If the relationship between the
input and the correct output is complicated and the network has enough hidden units to
model it accurately, there will typically be many different settings of the weights that
can model the training set almost perfectly, especially if there is only a limited amount of
labeled training data.  Each of these weight vectors will make different predictions on
held-out test data and almost all of them will do worse on the test data than on the training
data because the feature detectors have been tuned to work well together on the training
data but not on the test data. 

Overfitting can be reduced by using ``dropout'' to prevent complex co-adaptations on the
training data. On each presentation of each training case, each hidden unit is randomly
omitted from the network with a probability of 0.5, so a hidden unit cannot rely on other
hidden units being present. Another way to view the dropout procedure is as a very
efficient way of performing model averaging with neural networks. A good way to reduce the
error on the test set is to average the predictions produced by a very large number of
different networks.  The standard way to do this is to train many separate networks and
then to apply each of these networks to the test data, but this is computationally
expensive during both training and testing.  Random dropout makes it possible to train a
huge number of different networks in a reasonable time. There is almost certainly a
different network for each presentation of each training case but all of these networks
share the same weights for the hidden units that are present.

We use the standard, stochastic gradient descent procedure for training the dropout neural
networks on mini-batches of training cases, but we modify the penalty term that
is normally used to prevent the weights from growing too large. Instead of penalizing the
squared length (L2 norm) of the whole weight vector, we set an upper bound on the L2 norm
of the incoming weight vector for each individual hidden unit. If a weight-update violates
this constraint, we renormalize the weights of the hidden unit by division. Using a
constraint rather than a penalty prevents weights from growing very large no matter how
large the proposed weight-update is. This makes it possible to start with a very large
learning rate which decays during learning, thus allowing a far more thorough search of
the weight-space than methods that start with small weights and use a small learning rate.

At test time, we use the ``mean network'' that contains all of the hidden units but with
their outgoing weights halved to compensate for the fact that twice as many of them are
active.  In practice, this gives very similar performance to averaging over a large number
of dropout networks. In networks with a single hidden layer of $N$ units and a ``softmax''
output layer for computing the probabilities of the class labels, using the mean network
is exactly equivalent to taking the geometric mean of the probability distributions over
labels predicted by all $2^N$ possible networks.  Assuming the dropout networks do not all
make identical predictions, the prediction of the mean network is guaranteed to assign a
higher log probability to the correct answer than the mean of the log probabilities
assigned by the individual dropout networks \cite{HintonCD}.  Similarly, for regression
with linear output units, the squared error of the mean network is always better than the
average of the squared errors of the dropout networks.

\begin{figure}[t!]
\centerline{\includegraphics[scale=0.4]{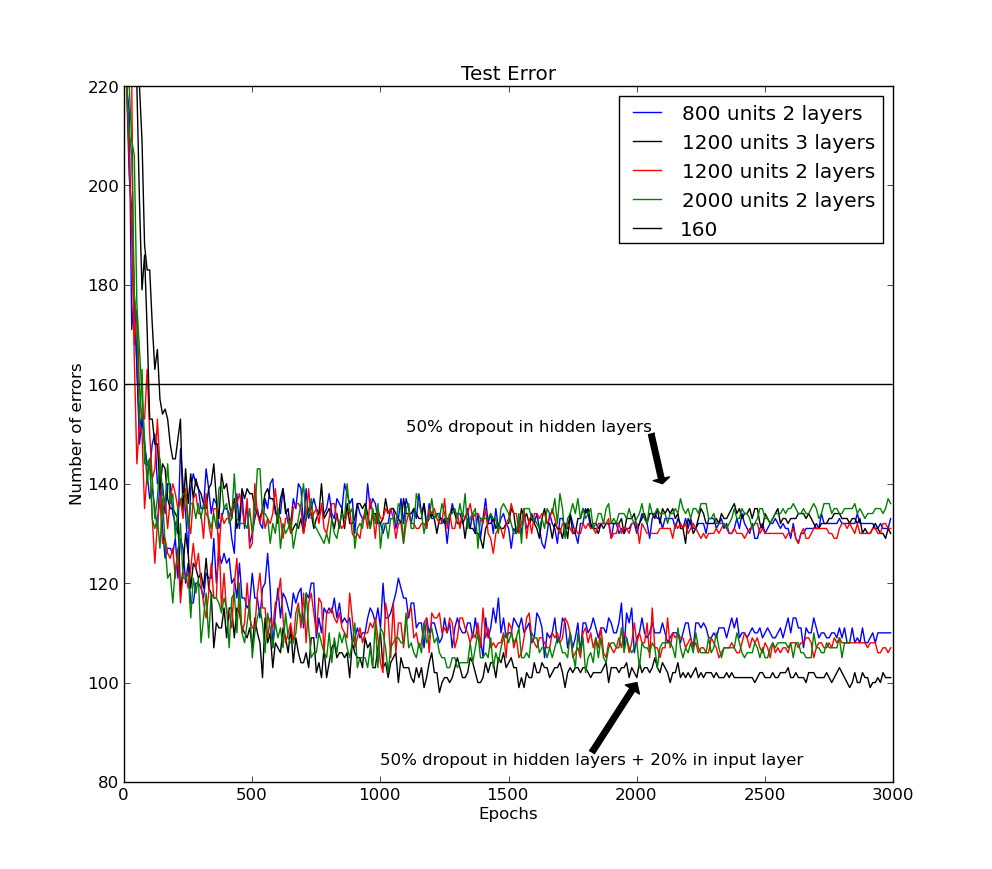}}
\caption{The error rate on the MNIST test set for a variety of neural network architectures
  trained with backpropagation using 50\% dropout for all hidden layers. The lower set of
  lines also use 20\% dropout
  for the input layer. The best previously published result for this task using
  backpropagation without pre-training or weight-sharing or enhancements of the training
  set is shown as a horizontal line. 
}
\label{fig:mnistplot}
\end{figure}

We initially explored the effectiveness of dropout using MNIST, a widely used benchmark
for machine learning algorithms. It contains 60,000 28x28 training images of individual
hand written digits and 10,000 test images. Performance on the test set can be greatly
improved by enhancing the training data with transformed images \cite{Ciresan2011} or by
wiring knowledge about spatial transformations into a convolutional neural network
\cite{Lecun} or by using generative pre-training to extract useful features from the
training images without using the labels \cite{Science}. Without using any of these
tricks, the best published result for a standard feedforward neural network is 160 errors
on the test set. This can be reduced to about 130 errors by using 50\% dropout with
separate L2 constraints on the incoming weights of each hidden unit and further reduced to
about 110 errors by also dropping out a random 20\% of the pixels (see figure~\ref{fig:mnistplot}).

Dropout can also be combined with generative pre-training, but in this case we use a small
learning rate and no weight constraints to avoid losing the feature detectors discovered
by the pre-training. The publically available, pre-trained deep belief net described in
\cite{Science} got 118 errors when it was fine-tuned using standard back-propagation and
92 errors when fine-tuned using 50\% dropout of the hidden units.  When the publically
available code at URL was used to pre-train a deep Boltzmann machine five times, the
unrolled network got 103, 97, 94, 93 and 88 errors when fine-tuned using standard
backpropagation and 83, 79, 78, 78 and 77 errors when using 50\% dropout of the hidden
units. The mean of 79 errors is a record for methods that do not use prior knowledge or
enhanced training sets (For details see Appendix~\ref{mnistsom}).

We then applied dropout to TIMIT, a widely used benchmark for recognition of clean speech
with a small vocabulary. Speech recognition systems use hidden Markov models (HMMs) to
deal with temporal variability and they need an acoustic model that determines how well a
frame of coefficients extracted from the acoustic input fits each possible state of each
hidden Markov model. Recently, deep, pre-trained, feedforward neural networks that map a
short sequence of frames into a probability distribution over HMM states have been shown
to outperform tradional Gaussian mixture models on both TIMIT \cite{MohamedIEEE} and a
variety of more realistic large vocabulary tasks \cite{DahlIEEE,JaitlyTR}.

\begin{figure}[t!]
\centerline{\includegraphics[scale = 0.40]{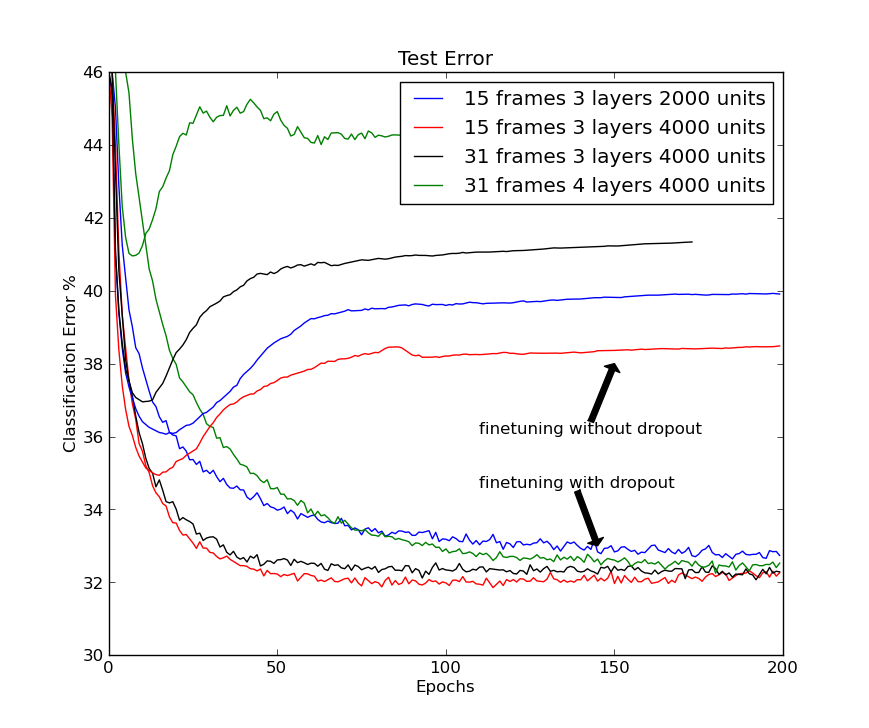}}
\caption{The frame {\it classification} error rate on the core test set of the TIMIT
  benchmark. Comparison of standard and dropout finetuning for different network
  architectures. Dropout of 50\% of the hidden 
  units and 20\% of the input units improves classification.}
\label{fig:timitplot}
\end{figure}

Figure~\ref{fig:timitplot} shows the frame {\it classification} error rate on the core test set of the
TIMIT benchmark when the central frame of a window is classified as belonging to the HMM
state that is given the highest probability by the neural net.  The input to the net is
21 adjacent frames with an advance of 10ms per frame. The neural net has 4 fully-connected
hidden layers of 4000 units per layer and 185 ``softmax'' output units that are subsequently
merged into the 39 distinct classes used for the benchmark. Dropout of 50\% of the hidden
units significantly improves classification for a variety of different network
architectures (see figure~\ref{fig:timitplot}).  To get the frame {\it recognition} rate, the class
probabilities that the neural network outputs for each frame are given to a decoder which
knows about transition probabilities between HMM states and runs the Viterbi algorithm to
infer the single best sequence of HMM states. Without dropout, the recognition rate is
$22.7$\% and with dropout this improves to $19.7$\%, which is a record for methods that do not use
any information about speaker identity.

\begin{figure}[t]
\centerline{\includegraphics[scale=0.4]{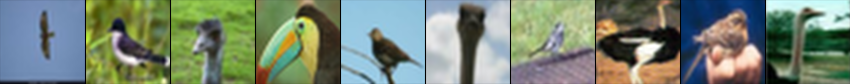}}
\caption{Ten examples of the class ``bird'' from the CIFAR-10 test set illustrating the
  variety of types of bird, viewpoint, lighting and background. The neural net gets all
  but the last two examples correct.}
\label{fig:cifar}
\end{figure}

CIFAR-10 is a benchmark task for object recognition. It uses 32x32 downsampled color
images of 10 different object classes that were found by searching the web for the names
of the class (e.g. dog) or its subclasses (e.g. Golden Retriever). These images were
labeled by hand to produce 50,000 training images and 10,000 test images in which there is
a single dominant object that could plausibly be given the class name \cite{kriz} (see
figure \ref{fig:cifar}). The best published error rate on the test set, without using
transformed data, is 18.5\% \cite{Coates2011}. We achieved an error rate of 16.6\% by using a
neural network with three convolutional hidden layers interleaved with three
``max-pooling'' layers that report the maximum activity in local pools of convolutional
units. These six layers were followed by one locally-connected layer (For details see Appendix~\ref{cifarsom}) . Using
dropout in the last hidden layer gives an error rate of 15.6\%.

ImageNet is an extremely challenging object recognition dataset consisting of thousands of
high-resolution images of thousands of classes of object \cite{Imagenet}. In 2010, a
subset of 1000 classes with roughly 1000 examples per class was the basis of an object
recognition competition in which the winning entry, which was actually an average of six
separate models, achieved an error rate of 47.2\% on the test set. The current state-of-the-art 
result on this dataset is 45.7\% \cite{winner45}. We achieved comparable performance of 
48.6\% error using a single neural
network with five convolutional hidden layers interleaved with ``max-pooling''
layer followed by two globally connected layers and a final
1000-way softmax layer. All layers had L2 weight constraints on the incoming weights of
each hidden unit. Using 50\% dropout in the sixth hidden layer reduces this to a record
42.4\% (For details see Appendix~\ref{imagenetsom}).

\begin{figure}[b]
\centerline{\includegraphics[scale=0.4]{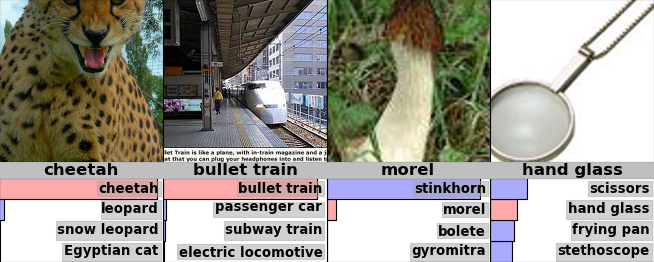}}
\caption{Some Imagenet test cases with the probabilities of the best 5 labels
  underneath. Many of the top 5 labels are quite plausible.}
\label{fig:imagenet}
\end{figure}

For the speech recognition dataset and both of the object recognition datasets it is
necessary to make a large number of decisions in designing the architecture of the net. We
made these decisions by holding out a separate validation set that was used to evaluate
the performance of a large number of different architectures and we then used the
architecture that performed best with dropout on the validation set to assess the
performance of dropout on the real test set. 

The Reuters dataset contains documents that have been labeled with a hierarchy of 
classes. We created training and test sets each containing 201,369 documents from 50 mutually
exclusive classes. Each document was represented by a vector of counts for 2000 common
non-stop words, with each count $C$ being transformed to $\log(1+C)$. A feedforward neural
network with 2 fully connected layers of 2000 hidden units trained with backpropagation
gets 31.05\% error on the test set. This is reduced to 29.62\% by using 50\% dropout (Appendix~\ref{reuterssom}).

We have tried various dropout probabilities and almost all of them improve the
generalization performance of the network.  For fully connected layers, dropout in all
hidden layers works better than dropout in only one hidden layer and more extreme
probabilities tend to be worse, which is why we have used 0.5 throughout this paper.  For
the inputs, dropout can also help, though it it often better to retain more than 50\% of
the inputs. It is also possible to adapt the individual dropout probability of each hidden
or input unit by comparing the average performance on a validation set with the average
performance when the unit is present. This makes the method work slightly
better. For datasets in which the required input-output mapping has a
number of fairly different regimes, performance can probably be further improved by making
the dropout probabilities be a learned function of the input, thus creating a
statistically efficient ``mixture of experts'' \cite{Jacobs} in which there are
combinatorially many experts, but each parameter gets adapted on a large fraction of the
training data.

Dropout is considerably simpler to implement than Bayesian model averaging which weights
each model by its posterior probability given the training data. For complicated model
classes, like feedforward neural networks, Bayesian methods typically use a Markov chain
Monte Carlo method to sample models from the posterior distribution \cite{NealThesis}.  By
contrast, dropout with a probability of $0.5$ assumes that all the models will eventually
be given equal importance in the combination but the learning of the shared weights takes
this into account. At test time, the fact that the dropout decisions are independent for
each unit makes it very easy to approximate the combined opinions of exponentially many
dropout nets by using a single pass through the mean net. This is far more efficient than
averaging the predictions of many separate models.

A popular alternative to Bayesian model averaging is ``bagging'' in which different models
are trained on different random selections of cases from the training set and all models
are given equal weight in the combination \cite{Breiman}. Bagging is most often used with
models such as decision trees because these are very quick to fit to data and very quick
at test time\cite{RandomForests}. Dropout allows a similar approach to be applied to feedforward
neural networks which are much more powerful models.  Dropout can be seen as an extreme
form of bagging in which each model is trained on a single case and each parameter of the
model is very strongly regularized by sharing it with the corresponding parameter in all
the other models. This is a much better regularizer than the standard method of shrinking
parameters towards zero.

A familiar and extreme case of dropout is ``naive bayes'' in which each input feature is
trained separately to predict the class label and then the predictive distributions of all
the features are multiplied together at test time. When there is very little training
data, this often works much better than logistic classification which trains each input
feature to work well in the context of all the other features.

Finally, there is an intriguing similarity between dropout and a recent theory
of the role of sex in evolution~\cite{Pap}.  One possible interpretation of the
theory of mixability articulated in~\cite{Pap} is that sex breaks up sets of
co-adapted genes and this means that achieving a function by using a large set
of co-adapted genes is not nearly as robust as achieving the same function,
perhaps less than optimally, in multiple alternative ways, each of which only
uses a small number of co-adapted genes. This allows evolution to avoid
dead-ends in which improvements in fitness require co- ordinated changes to a
large number of co-adapted genes. It also reduces the probability that small
changes in the environment will cause large decreases in fitness – a phenomenon
which is known as ``overfitting" in the field of machine learning.

\bibliography{dropout}
\bibliographystyle{Science}

\begin{scilastnote}
\item We thank N. Jaitly for help with TIMIT, H. Larochelle, R. Neal, K. Swersky and C.K.I. Williams for
helpful discussions, and NSERC, Google and Microsoft Research for
funding. GEH and RRS are members of the Canadian Institute for Advanced Research.
\end{scilastnote}
\appendix
\section{Experiments on MNIST}
\label{mnistsom}
\subsection{Details for dropout training}
\label{sec:mnist_dropout_nopre}
The MNIST dataset consists of 28 $\times$ 28 digit images - 60,000 for training and 10,000 for testing.
The objective is to classify the digit images into their correct digit class.
We experimented with neural nets of different architectures (different number of hidden units and
layers) to evaluate the sensitivity of the dropout method to these choices. We show
results for 4 nets (784-800-800-10, 784-1200-1200-10, 784-2000-2000-10,
784-1200-1200-1200-10). For each of these architectures we use the same dropout
rates - 50\% dropout for all hidden units and 20\% dropout for visible units.
We use stochastic gradient descent with 100-sized minibatches and a cross-entropy objective function.
An exponentially decaying learning rate is used 
that starts at the value of 10.0 (applied to the average gradient in each
minibatch). The learning rate is multiplied by 0.998 after each epoch of
training. The incoming weight vector corresponding to each hidden unit is constrained to have a
maximum squared length of $l$. If, as a result of an update, the squared length
exceeds $l$, the vector is scaled down so as to make it have a squared length of
$l$. Using cross validation we found that $l = 15$ gave best results.
Weights are initialzed to small random values drawn from a zero-mean normal distribution with
standard deviation 0.01. Momentum is used to speed up learning. The momentum
starts off at a value of 0.5 and is increased linearly to 0.99 over the first
500 epochs, after which it stays at 0.99. Also, the learning rate is multiplied
by a factor of (1-momentum). No weight decay is used.
Weights were updated at the end of each minibatch.
Training was done for 3000 epochs. 
The weight update takes the following form:
\begin{eqnarray*}
\Delta w^{t} & = & p^{t}\Delta w^{t-1} - (1-p^{t})\epsilon^{t}\langle \nabla_w L \rangle \\
w^{t} & = & w^{t-1} + \Delta w^{t},
\end{eqnarray*}
where,
\begin{eqnarray*}
\epsilon^{t} & = & \epsilon_{0}f^{t}\\
p^{t} &=& \begin{cases} 
            \frac{t}{T}p_{i} + (1-\frac{t}{T})p_{f} & t < T \\
            p_{f} & t \geq T
        \end{cases}
\end{eqnarray*}
with $\epsilon_0 = 10.0$, $f = 0.998$, $p_{i} = 0.5$, $p_{f} = 0.99$, $T = 500$.
While using a constant learning rate also gives improvements over standard
backpropagation, starting with a high learning rate and decaying it provided a
significant boost in performance. Constraining input vectors to have a fixed
length prevents weights from increasing arbitrarily in magnitude
irrespective of the learning rate. This gives the network a lot of opportunity
to search for a good configuration in the weight space. As the learning rate
decays, the algorithm is able to take smaller steps and finds the right step size
at which it can make learning progress. Using a high final momentum distributes
gradient information over a large number of updates making learning
stable in this scenario where each gradient computation is for a different
stochastic network.

\subsection{Details for dropout finetuning}
\label{sec:mnist_dropout_pre}
Apart from training a neural network starting from random weights, dropout can
also be used to finetune pretrained models. We found that finetuning a
model using dropout with a small learning rate can give much better performace
than standard backpropagation finetuning.

\emph{Deep Belief Nets} - We took a neural network pretrained using a Deep Belief Network~\cite{Science}.
It had a 784-500-500-2000 architecture and was trained using greedy layer-wise
Contrastive Divergence learning
\footnote{For code see \url{http://www.cs.toronto.edu/~hinton/MatlabForSciencePaper.html } }. Instead of fine-tuning it
with the usual backpropagation algorithm, we used the dropout version of it.
Dropout rate was same as before : 50\% for hidden units and 20\% for
visible units. A constant small learning rate of 1.0 was used. No constraint was imposed on
the length of incoming weight vectors. No weight decay was used. All other
hyper-parameters were set to be the same as before. The model was trained for 1000 epochs
with stochstic gradient descent using minibatches of size 100. While standard
back propagation gave about 118 errors, dropout decreased the errors to about 92.

\emph{Deep Boltzmann Machines} - We also took a pretrained Deep Boltzmann
Machine \cite{DBM} \footnote{ For code see
\url{http://www.utstat.toronto.edu/~rsalakhu/DBM.html }}
(784-500-1000-10) and finetuned it using dropout-backpropagation. The
model uses a 1784 - 500 - 1000 - 10 architecture (The extra 1000 input
units come from the mean-field activations of the second layer of hidden units
in the DBM, See~\cite{DBM} for details).
All finetuning hyper-parameters were set to be the same as the ones used for a Deep Belief Network.
We were able to get a mean of about 79 errors with dropout whereas usual finetuning gives about 94 errors.

\begin{figure}[ht]
\centerline{
\subfloat[]{\includegraphics[width=0.68\linewidth]{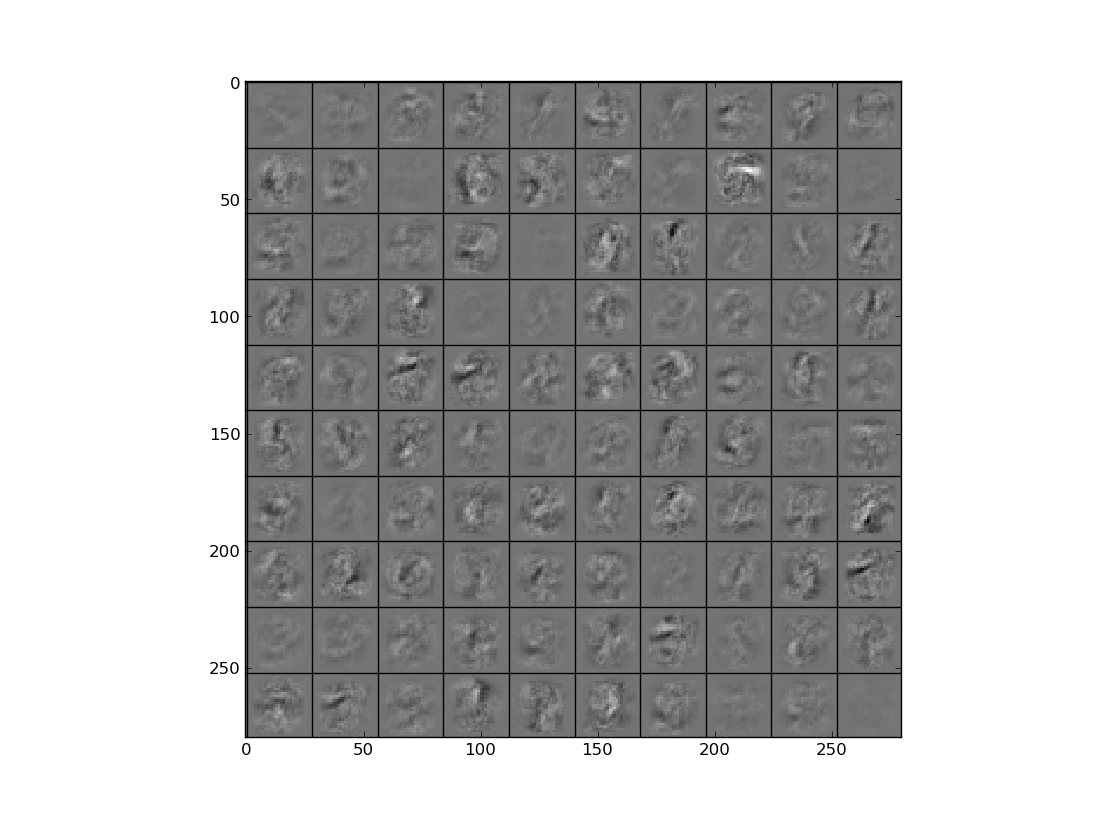}} 
\subfloat[]{\includegraphics[width=0.68\linewidth]{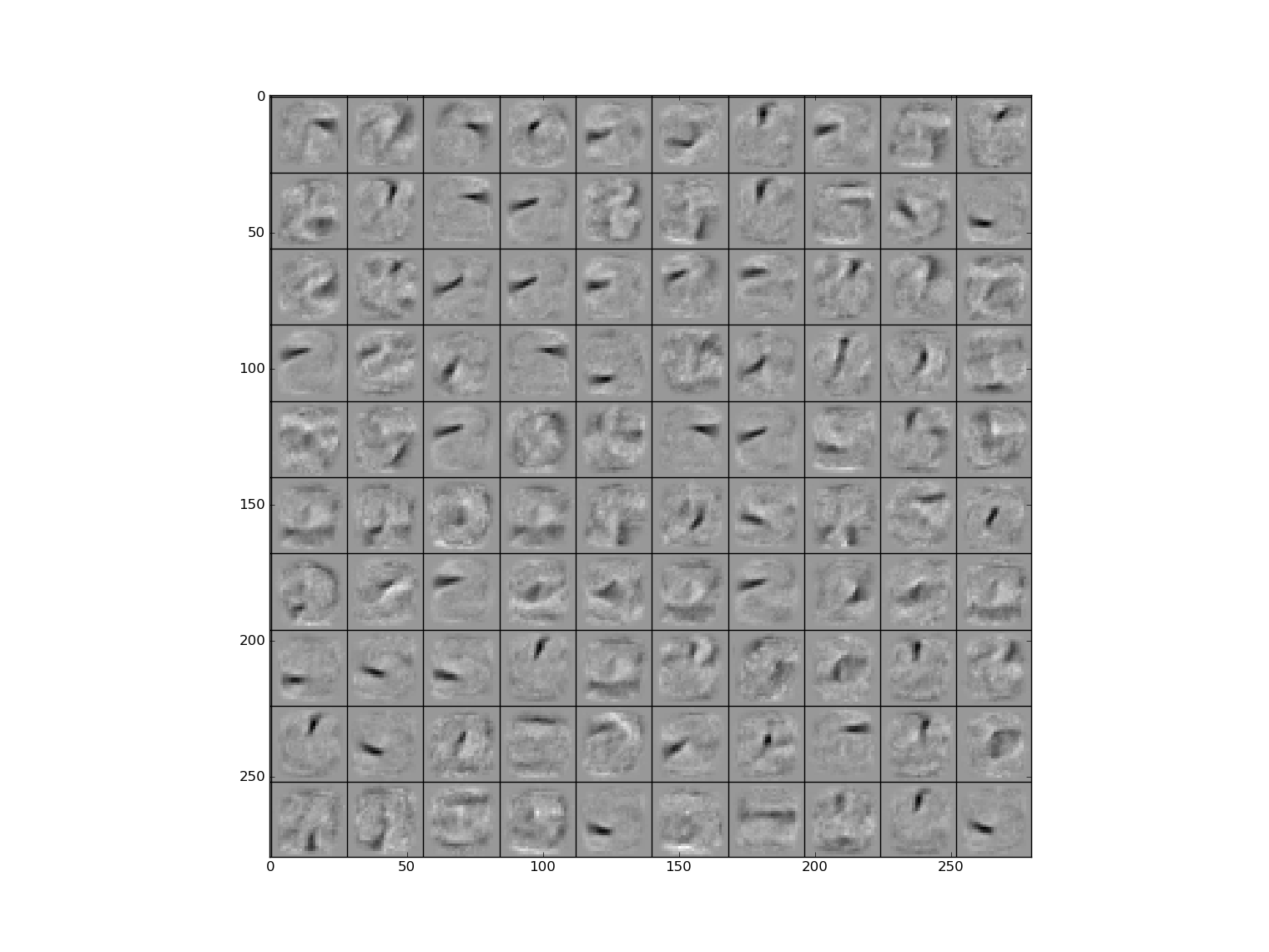}}
}
\caption{Visualization of features learned by first layer hidden units for (a)
backprop and (b) dropout on the MNIST dataset.}
\label{fig:features}
\end{figure}
\subsection{Effect on features}
One reason why dropout gives major improvements over backpropagation is that it encourages each individual
hidden unit to learn a useful feature without relying on specific other hidden units to
correct its mistakes.  In order to verify this and better understand the effect
of dropout on feature learning, we look at the first level of features learned by a 784-500-500
neural network without any generative pre-training. The features are shown in Figure~\ref{fig:features}. Each panel shows 100 random
features learned by each network. The features that dropout
learns are simpler and look like strokes, whereas the ones learned by 
standard backpropagation are difficult to interpret. This
confirms that dropout indeed forces the discriminative model to learn
good features which are less co-adapted and leads to better generalization.


\section{Experiments on TIMIT}
\label{timitsom}
The TIMIT Acoustic-Phonetic Continuous Speech Corpus is a standard dataset used
for evaluation of automatic speech recognition systems. It consists of
recordings of 630 speakers of 8 dialects of American English each
reading 10 phonetically-rich sentences. It also comes with the word and
phone-level transcriptions of the speech. The objective is to convert a given 
speech signal into a transcription sequence of phones.
This data needs to be pre-processed to extract input features and output targets.
We used Kaldi, an open source code library for speech
\footnote{\url{ http://kaldi.sourceforge.net }}, to
pre-process the dataset so that our results can be reproduced exactly.
The inputs to our networks are log filter bank responses. They are extracted
for 25 ms speech windows with strides of 10 ms. 

\begin{figure}[h!]
\centerline{
\subfloat[]{\includegraphics[scale=0.36]{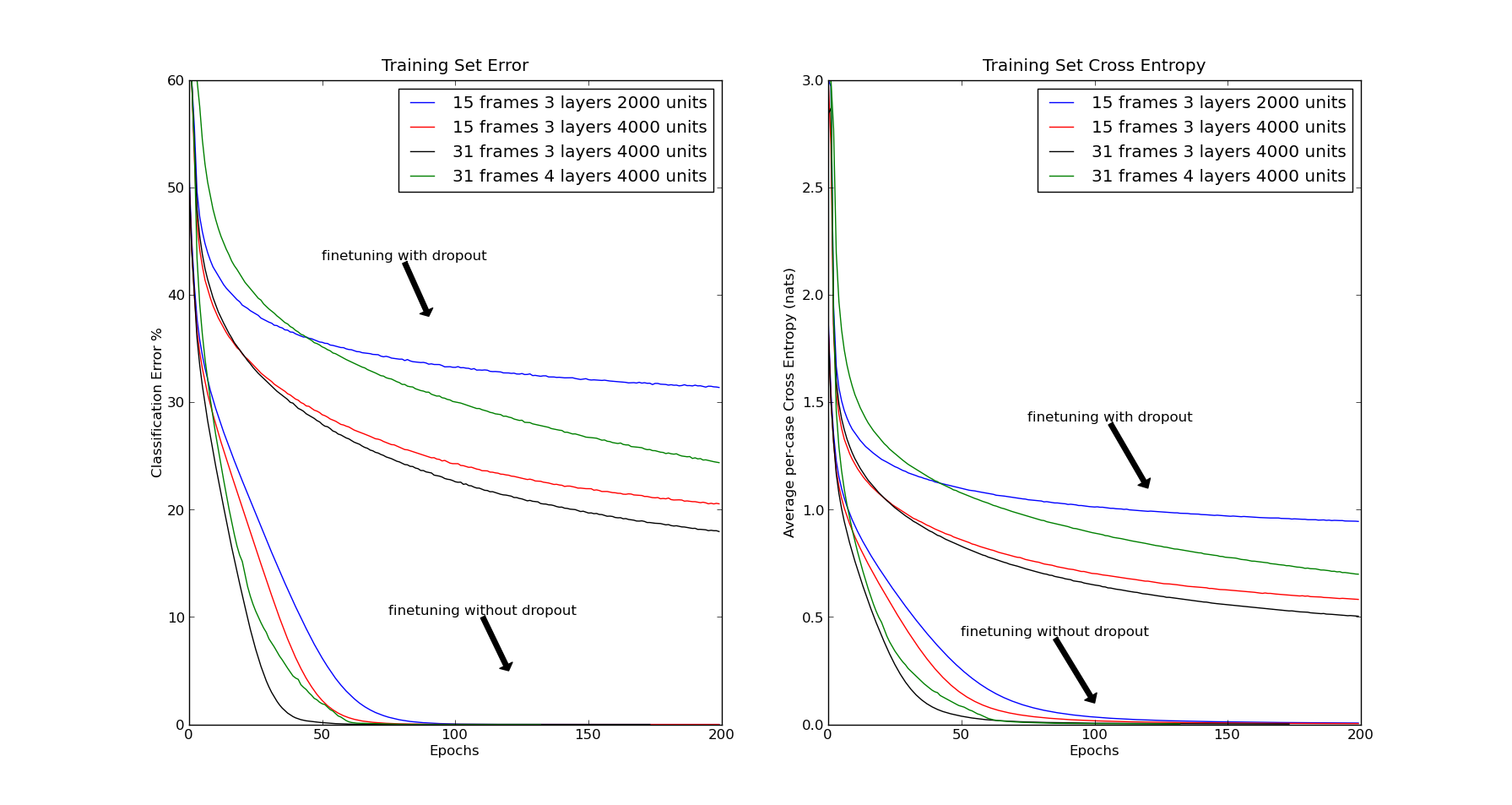}}}
\centerline{
\subfloat[]{\includegraphics[scale=0.36]{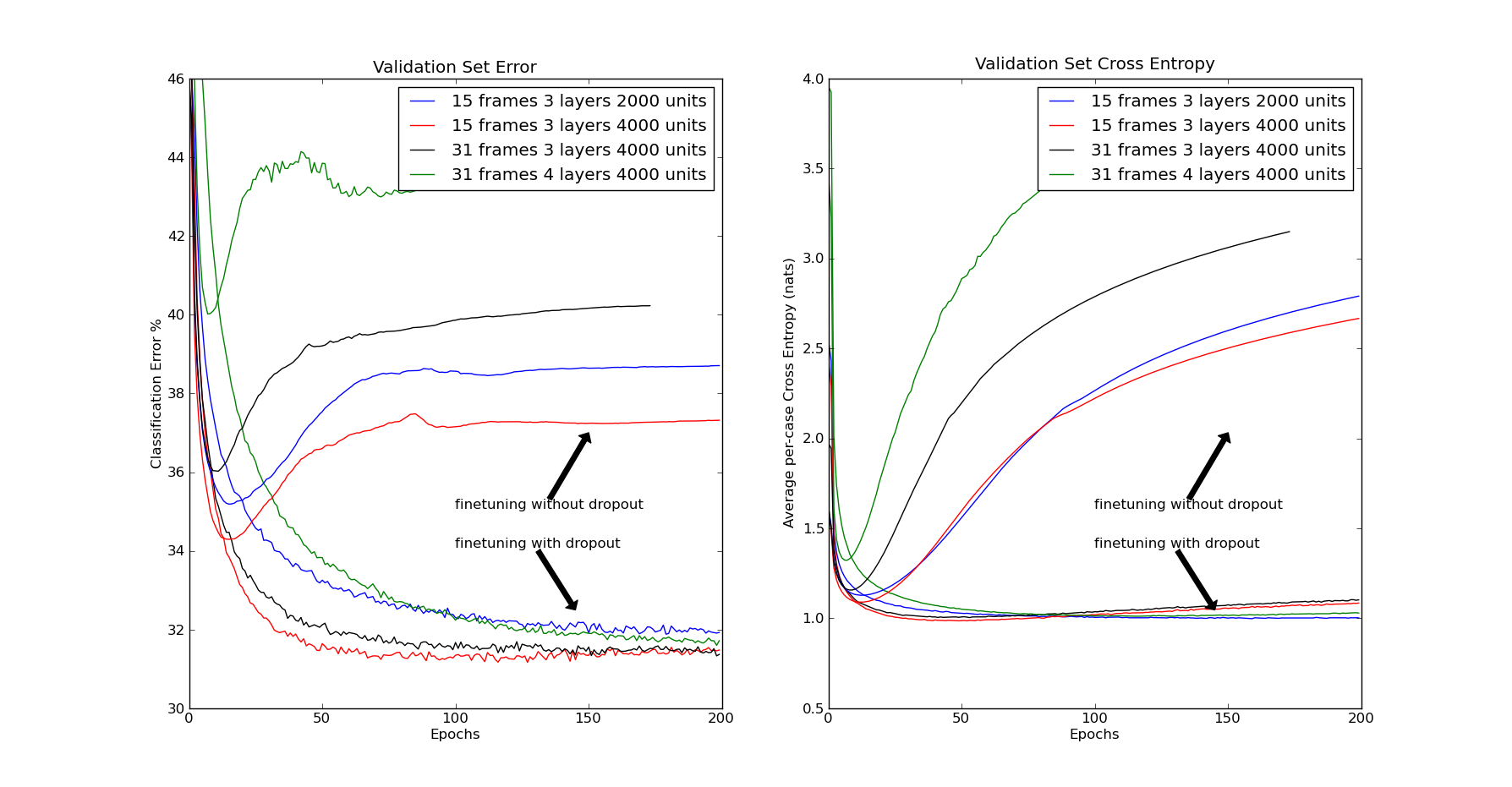}}
}
\caption{Frame classification error and cross-entropy on the (a) Training and
(b) Validation set as learning progresses. 
The training error is computed using the stochastic nets.}
\label{fig:timit}
\end{figure}

Each dimension of the input representation was normalized to have mean 0 and variance 1.
Minibatches of size 100 were used for both pretraining and dropout finetuning. We
tried several network architectures by varying the number of input frames (15 and 31), number of
layers in the neural network (3, 4 and 5) and the number of hidden units in each
layer (2000 and 4000). 
Figure~\ref{fig:timit} shows the validation
error curves for a number of these combinations. Using dropout
consistently leads to lower error rates. 

\subsection{Pretraining}
\label{sec:timit_pre}
For all our experiments on TIMIT, we pretrain the neural network with a
Deep Belief Network~\cite{Science}.
Since the inputs are real-valued, the first layer was pre-trained as a Gaussian RBM. Visible biases were initialized
to zero and weights to random numbers sampled from a zero-mean normal distribution with
standard deviation 0.01. The variance of each visible unit was set to 1.0 and
not learned. Learning was done by minimizing Contrastive Divergence. Momentum
was used to speed up learning. Momentum started at 0.5 and was increased
linearly to 0.9 over 20 epochs. A learning rate of 0.001 on the average gradient
was used (which was then multiplied by 1-momentum). An L2 weight decay of 0.001 was used. The model was
trained for 100 epochs.

All subsequent layers were trained as binary RBMs. A learning rate of 0.01 was
used. The visible bias of each unit was initialized to $\log(p/(1-p))$ where $p$
was the mean activation of that unit in the dataset. All other hyper-parameters
were set to be the same as those we used for the Gaussian RBM. Each layer was trained for 50 epochs.

\subsection{Dropout Finetuning}
\label{sec:timit_dropout}
The pretrained RBMs were used to initialize the weights in a neural network. The
network was then finetuned with dropout-backpropagation.
Momentum was increased from 0.5 to 0.9 linearly over 10 epochs. A small constant learning
rate of 1.0 was used (applied to the average gradient on a minibatch). All other hyperparameters
are the same as for MNIST dropout finetuning. The model needs to be run for
about 200 epochs to converge. The same network was also finetuned with standard
backpropagation using a smaller learning rate of 0.1, keeping all other
hyperparameters 

Figure~\ref{fig:timit} shows the frame classification error and cross-entropy
objective value on the training and validation sets. We compare the performance
of dropout and standard backpropagation on several network architectures and
input representations. Dropout consistently achieves lower error and
cross-entropy. It significantly controls overfitting, making the method robust
to choices of network architecture. It allows much larger nets to be trained and removes
the need for early stopping. We also observed that the final error obtained by
the model is not very sensitive to the choice of learning rate and momentum.

\section{Experiments on Reuters}
\label{reuterssom}

Reuters Corpus Volume I (RCV1-v2) \cite{Lewis2004} is an archive of 
804,414 newswire stories that have been manually categorized into 103 topics\footnote{The corpus is
available at http://www.ai.mit.edu/projects/jmlr/papers/volume5/lewis04a/lyrl2004\_rcv1v2\_README.htm}. 
The corpus 
covers four major groups: corporate/industrial, economics, government/social, and markets. 
Sample topics include Energy Markets, Accounts/Earnings, Government Borrowings, Disasters and 
Accidents, Interbank Markets, Legal/Judicial, Production/Services, etc. The topic classes form 
a tree which is typically of depth three.

We took the dataset and split it into 63 classes based on the 
the 63 categories at the second-level of the category tree.
We removed 11 categories that did not have any data and one category that
had only 4 training examples. We also removed one category that covered a huge
chunk (25\%) of the examples. This left us with 50 classes and 402,738 documents.
We divided the documents into equal-sized training and test sets randomly. Each
document was represented using the 2000 most frequent non-stopwords in the dataset.

\begin{figure}[h]
\subfloat[]{\includegraphics[scale=0.4]{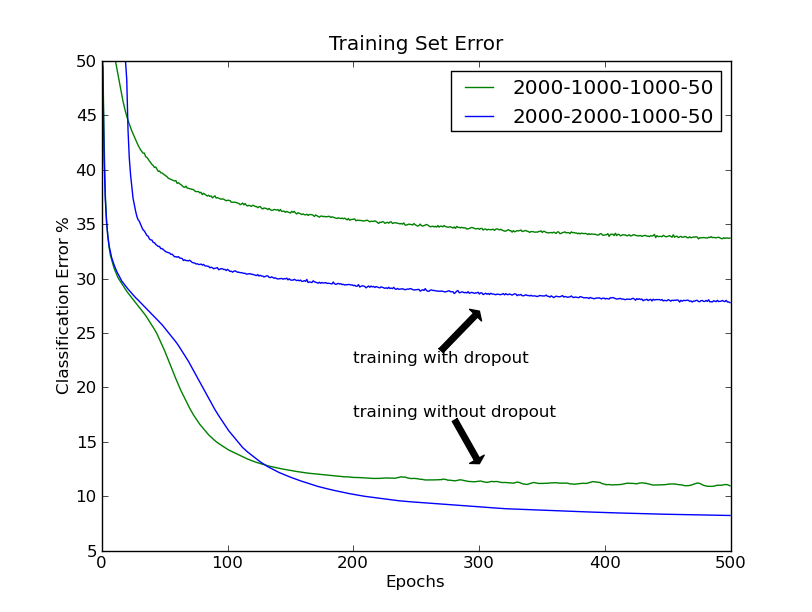}}
\subfloat[]{\includegraphics[scale=0.4]{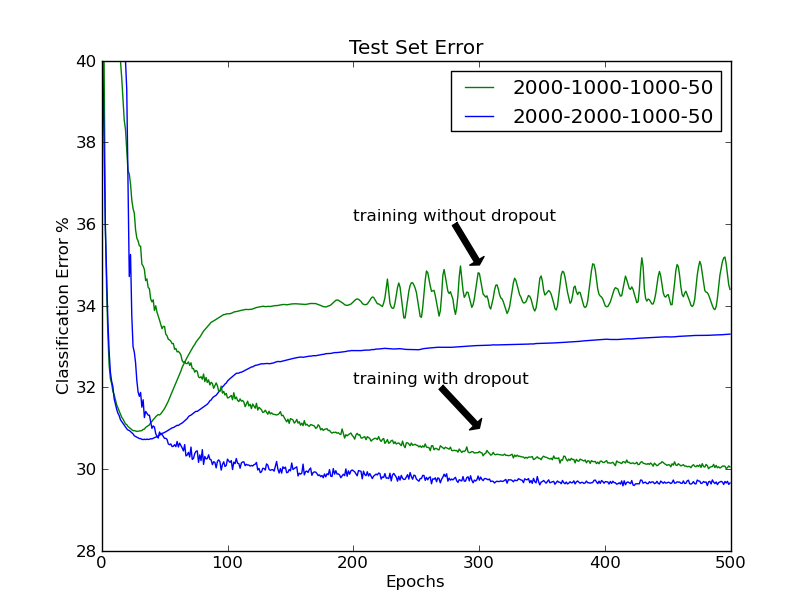}}
\caption{Classification error rate on the (a) training and (b) validation sets of the Reuters dataset as learning progresses.
The training error is computed using the stochastic nets.}
\label{fig:reuters}
\end{figure}

We trained a neural network using dropout-backpropagation and compared it with
standard backpropagation. We used a 2000-2000-1000-50 architecture. The training
hyperparameters are same as that in MNIST dropout training (Appendix~\ref{sec:mnist_dropout_nopre}). Training was done
for 500 epochs.

Figure~\ref{fig:reuters} shows the training and test set errors as learning
progresses. We show two nets - one with a 2000-2000-1000-50 and
another with a 2000-1000-1000-50 architecture trained with and without dropout.
As in all previous datasets discussed so far, we obtain significant improvements
here too. The learning not only results in better generalization, but also
proceeds smoothly, without the need for early stopping.

\section{Tiny Images and CIFAR-10}
\label{cifarsom}

The Tiny Images dataset contains 80 million $32\times32$ color images
collected from the web. The images were found by searching various
image search engines for English nouns, so each image comes with a
very unreliable label, which is the noun that was used to find it.
The CIFAR-10 dataset is a subset of the Tiny Images dataset which
contains 60000 images divided among ten classes\footnote{The CIFAR dataset is 
available at http://www.cs.toronto.edu/$\sim$kriz/cifar.html. }. Each class contains
5000 training images and 1000 testing images. The classes are \emph{airplane},
\emph{automobile}, \emph{bird}, \emph{cat}, \emph{deer}, \emph{dog},
\emph{frog}, \emph{horse}, \emph{ship}, and \emph{truck}. The CIFAR-10
dataset was obtained by filtering the Tiny Images dataset to remove
images with incorrect labels. The CIFAR-10 images are highly varied,
and there is no canonical viewpoint or scale at which the objects
appear. The only criteria for including an image were that the image
contain one dominant instance of a CIFAR-10 class, and that the object
in the image be easily identifiable as belonging to the class indicated
by the image label.

\section{ImageNet}
\label{imagenetsom}

ImageNet is a dataset of millions of labeled images in thousands of
categories. The images were collected from the web and labelled by
human labellers using Amazon's Mechanical Turk crowd-sourcing tool.
In 2010, a subset of roughly 1000 images in each of 1000 classes was
the basis of an object recognition competition, a part of the Pascal
Visual Object Challenge. This is the version of ImageNet on which
we performed our experiments. In all, there are roughly 1.3 million
training images, 50000 validation images, and 150000 testing images.
This dataset is similar in spirit to the CIFAR-10, but on a much bigger
scale. The images are full-resolution, and there are 1000 categories
instead of ten. Another difference is that the ImageNet images often
contain multiple instances of ImageNet objects, simply due to the
sheer number of object classes. For this reason, even a human would
have difficulty approaching perfect accuracy on this dataset. For
our experiments we resized all images to $256\times256$ pixels.

\section{Convolutional Neural Networks}

Our models for CIFAR-10 and ImageNet are deep, feed-forward convolutional
neural networks (CNNs). Feed-forward neural networks are models which
consist of several layers of {}``neurons'', where each neuron in
a given layer applies a linear filter to the outputs of the neurons
in the previous layer. Typically, a scalar bias is added to the filter
output and a nonlinear activation function is applied to the result
before the neuron's output is passed to the next layer. The linear
filters and biases are referred to as \emph{weights}, and these are
the parameters of the network that are learned from the training data.

CNNs differ from ordinary neural networks in several ways. First,
neurons in a CNN are organized topographically into a bank that reflects
the organization of dimensions in the input data. So for images, the
neurons are laid out on a 2D grid. Second, neurons in a CNN apply
filters which are local in extent and which are centered at the neuron's
location in the topographic organization. This is reasonable for datasets
where we expect the dependence of input dimensions to be a decreasing
function of distance, which is the case for pixels in natural images.
In particular, we expect that useful clues to the identity of the
object in an input image can be found by examining small local neighborhoods
of the image. Third, all neurons in a bank apply the same filter,
but as just mentioned, they apply it at different locations in the
input image. This is reasonable for datasets with roughly stationary
statistics, such as natural images. We expect that the same kinds
of structures can appear at all positions in an input image, so it
is reasonable to treat all positions equally by filtering them in
the same way. In this way, a bank of neurons in a CNN applies a convolution
operation to its input. A single layer in a CNN typically has multiple
banks of neurons, each performing a convolution with a different filter.
These banks of neurons become distinct input channels into the next
layer. The distance, in pixels, between the boundaries of the receptive
fields of neighboring neurons in a convolutional bank determines the
\emph{stride} with which the convolution operation is applied. Larger
strides imply fewer neurons per bank. Our models use a stride of one
pixel unless otherwise noted.

One important consequence of this convolutional shared-filter architecture
is a drastic reduction in the number of parameters relative to a neural
net in which all neurons apply different filters. This reduces the
net's representational capacity, but it also reduces its capacity
to overfit, so dropout is far less advantageous in convolutional layers.

\subsection{Pooling}

CNNs typically also feature {}``pooling'' layers which summarize
the activities of local patches of neurons in convolutional layers.
Essentially, a pooling layer takes as input the output of a convolutional
layer and subsamples it. A pooling layer consists of pooling units
which are laid out topographically and connected to a local neighborhood
of convolutional unit outputs from the same bank. Each pooling unit
then computes some function of the bank's output in that neighborhood.
Typical functions are maximum and average. Pooling layers with such
units are called max-pooling and average-pooling layers, respectively.
The pooling units are usually spaced at least several pixels apart,
so that there are fewer total pooling units than there are convolutional
unit outputs in the previous layer. Making this spacing smaller than
the size of the neighborhood that the pooling units summarize produces
\emph{overlapping pooling.} This variant makes the pooling
layer produce a coarse coding of the convolutional unit outputs, which
we have found to aid generalization in our experiments. We refer to
this spacing as the \emph{stride} between pooling units, analogously
to the stride between convolutional units. Pooling layers introduce
a level of local translation invariance to the network, which improves
generalization. They are the analogues of \emph{complex cells} in
the mammalian visual cortex, which pool activities of multiple simple
cells. These cells are known to exhibit similar phase-invariance properties.

\subsection{Local response normalization}

Our networks also include response normalization layers. This type
of layer encourages competition for large activations among neurons
belonging to different banks. In particular, the activity $a_{x,y}^{i}$
of a neuron in bank $i$ at position $(x,y)$ in the topographic organization
is divided by 
\[
\left(1+\alpha\sum_{j=i-N/2}^{i+N/2}(a_{x,y}^{j})^{2}\right)^{\beta}
\]
where the sum runs over $N$ {}``adjacent'' banks of neurons at
\emph{the same position in the topographic organization}. The
ordering of the banks is of course arbitrary and determined before
training begins. Response normalization layers implement a form of
lateral inhibition found in real neurons. The constants $N,\alpha$,
and $\beta$ are hyper-parameters whose values are determined using
a validation set.

\subsection{Neuron nonlinearities}

All of the neurons in our networks utilize the max-with-zero nonlinearity.
That is, their output is computed as $a_{x,y}^{i}=\max(0,z_{x,y}^{i})$
where $z_{x,y}^{i}$ is the total input to the neuron (equivalently,
the output of the neuron's linear filter added to the bias). This nonlinearity has several
advantages over traditional saturating neuron models, including a
significant reduction in the training time required to reach a given
error rate. This nonlinearity also reduces the need for contrast-normalization
and similar data pre-processing schemes, because neurons with this
nonlinearity do not saturate -- their activities simply scale up when
presented with unusually large input values. Consequently, the only
data pre-processing step which we take is to subtract the mean activity
from each pixel, so that the data is centered. So we train our networks
on the (centered) raw RGB values of the pixels.

\subsection{Objective function}

Our networks maximize the multinomial logistic regression objective,
which is equivalent to minimizing the average across training cases
of the cross-entropy between the true label distribution and the model's
predicted label distribution.

\subsection{Weight initialization}

We initialize the weights in our model from a zero-mean normal distribution
with a variance set high enough to produce positive inputs into the
neurons in each layer. This is a slightly tricky point when using
the max-with-zero nonlinearity. If the input to a neuron is always
negative, no learning will take place because its output will be uniformly
zero, as will the derivative of its output with respect to its input.
Therefore it's important to initialize the weights from a distribution
with a sufficiently large variance such that all neurons are likely
to get positive inputs at least occasionally. In practice, we simply
try different variances until we find an initialization that works.
It usually only takes a few attempts. We also find that initializing
the biases of the neurons in the hidden layers with some positive
constant (1 in our case) helps get learning off the ground, for the
same reason.

\subsection{Training}

We train our models using stochastic gradient descent with a batch
size of 128 examples and momentum of 0.9. Therefore the update rule
for weight $w$ is
\begin{eqnarray*}
v_{i+1} & = & 0.9 v_{i}+\epsilon<\frac{\partial E}{\partial w_{i}}>_{i}\\
w_{i+1} & = & w_{i}+v_{i+1}
\end{eqnarray*}
where $i$ is the iteration index, $v$ is a momentum variable, $\epsilon$
is the learning rate, and $<\frac{\partial E}{\partial w_{i}}>_{i}$
is the average over the $i^{th}$ batch of the derivative of the objective
with respect to $w_{i}$. We use the publicly available \texttt{cuda-convnet}
package to train all of our models on a single NVIDIA GTX 580 GPU.
Training on CIFAR-10 takes roughly 90 minutes. Training on ImageNet
takes roughly four days with dropout and two days without.

\subsection{Learning rates}

We use an equal learning rate for each layer, whose value we determine
heuristically as the largest power of ten that produces reductions
in the objective function. In practice it is typically of the order
$10^{-2}$ or $10^{-3}$. We reduce the learning rate twice by a
factor of ten shortly before terminating training.

\section{Models for CIFAR-10}

Our model for CIFAR-10 without dropout is a CNN with three convolutional
layers. Pooling layers follow all three. All of the pooling layers
summarize a $3\times3$ neighborhood and use a stride of 2. The pooling
layer which follows the first convolutional layer performs max-pooling,
while the remaining pooling layers perform average-pooling. Response
normalization layers follow the first two pooling layers, with $N=9$,
$\alpha=0.001$, and $\beta=0.75$. The upper-most pooling layer is
connected to a ten-unit softmax layer which outputs a probability
distribution over class labels. All convolutional layers have 64 filter banks
and use a filter size of $5\times5$ (times the number of channels
in the preceding layer).

Our model for CIFAR-10 with dropout is similar, but because dropout
imposes a strong regularization on the network, we are able to use
more parameters. Therefore we add a fourth weight layer, which takes
its input from the third pooling layer. This weight layer is \emph{locally-connected
but not convolutional.} It is like a convolutional layer in which
filters in the same bank do not share weights. This layer contains
16 banks of filters of size $3\times3$. This is the layer in which
we use 50\% dropout. The softmax layer takes its input from this fourth
weight layer.

\section{Models for ImageNet}

Our model for ImageNet with dropout is a CNN which is trained on $224\times224$
patches randomly extracted from the $256\times256$ images, as well
as their horizontal reflections. This is a form of data augmentation
that reduces the network's capacity to overfit the training data and
helps generalization. The network contains seven weight layers. The
first five are convolutional, while the last two are globally-connected.
Max-pooling layers follow the first, second, and fifth convolutional
layers. All of the pooling layers summarize a $3\times3$ neighborhood
and use a stride of 2. Response-normalization layers follow the first
and second pooling layers. The first convolutional layer has 64 filter
banks with $11\times11$ filters which it applies with a stride of
4 pixels (this is the distance between neighboring neurons in a bank).
The second convolutional layer has 256 filter banks with $5\times5$
filters. This layer takes two inputs. The first input to this layer
is the (pooled and response-normalized) output of the first convolutional
layer. The 256 banks in this layer are divided arbitrarily into groups
of 64, and each group connects to a unique random 16 channels from
the first convolutional layer. The second input to this layer is a
subsampled version of the original image ($56\times56$), which is
filtered by this layer with a stride of 2 pixels. The two maps resulting
from filtering the two inputs are summed element-wise (they have exactly
the same dimensions) and a max-with-zero nonlinearity is applied to
the sum in the usual way. The third, fourth, and fifth convolutional
layers are connected to one another without any intervening pooling
or normalization layers, but the max-with-zero nonlinearity is applied
at each layer after linear filtering. The third convolutional layer
has 512 filter banks divided into groups of 32, each group connecting
to a unique random subset of 16 channels produced by the (pooled, normalized)
outputs of the second convolutional layer. The fourth and fifth convolutional
layers similarly have 512 filter banks divided into groups of 32,
each group connecting to a unique random subset of 32 channels produced
by the layer below. The next two weight layers are globally-connected,
with 4096 neurons each. In these last two layers we use 50\% dropout.
Finally, the output of the last globally-connected layer is fed to
a 1000-way softmax which produces a distribution over the 1000 class
labels. We test our model by averaging the prediction of the net on
ten $224\times224$ patches of the $256\times256$ input image: the
center patch, the four corner patches, and their horizontal reflections.
Even though we make ten passes of each image at test time, we are
able to run our system in real-time.

Our model for ImageNet without dropout is similar, but without the
two globally-connected layers which create serious overfitting when
used without dropout.

In order to achieve state-of-the-art performance on the validation
set, we found it necessary to use the very complicated network
architecture described above. Fortunately, the complexity of this
architecture is not the main point of our paper. What we wanted to
demonstrate is that dropout is a significant help even for the very complex
neural nets that have been developed by the joint efforts of many
groups over many years to be really good at object recognition. This
is clearly demonstrated by the fact that using non-convolutional
higher layers with a lot of parameters leads to a big improvement with
dropout but makes things worse without dropout.

\end{document}